\documentclass{article}

\usepackage{times}
\usepackage{graphicx} % more modern
\usepackage{subfigure} 

\usepackage{natbib}

\usepackage{algorithm}
\usepackage{algorithmic}

\usepackage{kotex}
\usepackage{amsthm}
\usepackage{amsmath}

\theoremstyle{definition}
\newtheorem{definition}{Definition}[section]
\usepackage{amssymb}
\usepackage{float}

\usepackage{hyperref}

\usepackage[accepted]{icml2017}

\icmltitlerunning{Micro-Objective Learning : Accelerating Deep Reinforcement Learning through the Discovery of Continuous Subgoals}

\begin{document} 

\twocolumn[
\icmltitle{Micro-Objective Learning : Accelerating Deep Reinforcement Learning through the Discovery of Continuous Subgoals}



% you can specify symbols, otherwise they are numbered in order

\begin{icmlauthorlist}
\icmlauthor{Sungtae Lee}{com,med}
\icmlauthor{Sang-Woo Lee}{com}
\icmlauthor{Jinyoung Choi}{cog}
\icmlauthor{Dong-Hyun Kwak}{neuro}
\icmlauthor{Byoung-Tak Zhang}{com,cog,neuro}
\end{icmlauthorlist}

\icmlaffiliation{com}{School of Computer Science and Engineering, Seoul National University}
\icmlaffiliation{med}{College of Medicine, Yonsei University}
\icmlaffiliation{cog}{Interdisciplinary Program in Cognitive Science, Seoul National University}
\icmlaffiliation{neuro}{Interdisciplinary Program in Neuroscience, Seoul National University}

\icmlcorrespondingauthor{Byoung-Tak Zhang}{btzhang@bi.snu.ac.kr}

\icmlkeywords{subgoal, reinforcement learning, ICML}

\vskip 0.3in
]



\printAffiliationsAndNotice{}  % leave blank if no need to mention equal contribution
%\footnotetext{hi}

\begin{abstract}
Recently, reinforcement learning has been successfully applied to the logical game of Go, various Atari games, and even a 3D game, Labyrinth, though it continues to have problems in sparse reward settings. It is difficult to explore, but also difficult to exploit, a small number of successes when learning policy. To solve this issue, the subgoal and option framework have been proposed. However, discovering subgoals online is too expensive to be used to learn options in large state spaces. We propose Micro-objective learning (MOL) to solve this problem. The main idea is to estimate how important a state is while training and to give an additional reward proportional to its importance.  We evaluated our algorithm in two Atari games: Montezuma’s Revenge and Seaquest. With three experiments to each game, MOL significantly improved the baseline scores. Especially in Montezuma’s Revenge, MOL achieved two times better results than the previous state-of-the-art model.
\end{abstract} 

\section{Introduction}
\label{introduction}
Recent advances in deep reinforcement learning\cite{mnih2013playing, van2016deep, wang2015dueling} have triggered subgoal and option research within reinforcement learning using deep neural networks. Here we investigate Micro-Objective Learning, a new technique to discover important states for a complex task and exploit them to accelerate learning.

Many techniques have been investigated with respect to subgoals. \cite{kulkarni2016deep} uses a Successor map, which approximates the count of the successive states given (state, action) pairs. It successfully extracted subgoals in the simple grid world and 3D Doom task. However, it did not exploit the subgoals while discovering them online to accelerate learning. \cite{kulkarni2016hierarchical} used heuristically defined subgoals and two neural networks: One to maximize the external reward, and the other to maximize the internal reward to achieve subgoals. This \cite{kulkarni2016hierarchical} paper has had promising results in Montezuma’s Revenge, an Atari game notorious for its difficult exploration. \cite{bacon2016option} has incorporated option learning into policy gradient theorem to directly do end-to-end learning of the options for semi-MDP (Markov Decision Process) and has been tested on four Atari games\cite{bellemare2013arcade} but left the number of options as a hyper-parameter.

Despite the promising results by \cite{kulkarni2016hierarchical, bacon2016option}, several drawbacks still exist. 1) Although subgoals and options are inseparable, discovering the subgoals online while learning options has been extremely difficult as they need to be learned in addition to original reinforcement learning algorithms. As far as we know, online learning of subgoals while exploiting them to accelerate learning in large state-space, such as the Atari domain, has not been accomplished. In \cite{bacon2016option}, they did not use the idea of subgoals to learn options.

Additionally, 2) the definition of subgoal has been somewhat ambiguously defined. In \cite{murata2007introduction}, a subgoal was defined as a set of states that are known to be visited when reaching the goal. However, every state can be considered a subgoal. For example, when a robot is learning to move from one room to another and a door is in between, the door can be viewed as a subgoal in a cognitive way. Also, we can consider a desk or a chair that one can pass (but does not need to pass) to get to the other room as a subgoal with lower importance than the door.

In this paper, we present a hierarchical reinforcement learning algorithm, Micro-Objective Learning(MOL). We first define micro-objectives, a set of states with a continuous measurement of importance assigned to each element, which is a micro-version of the subgoal. We induce the logic that frequently visited states can be considered as subgoals, from \cite{mcgovern2001accelerating}, however only successful trajectories are used to estimate the importance of micro-objectives. Similar to the empirical results in \cite{mcgovern2001accelerating}, simply counting all states results in distracted subgoals, while counting the states only when visited for the first time gives clearer results.

There are drawbacks of first-visit counting as mentioned in \cite{mcgovern2001accelerating}: It is noisy, and impractical in large or continuous state-space domains. Though it appears noisy from the traditional subgoal point of view, it does help to approximate the importance of micro-objective states. Also, noisy subgoals are normally discouraged because learning option policy takes time. However, since we give an additional reward to the original MDP rather than explicitly creating macro-actions, estimating the importance does not need to be precise. Additionally, we use recent pixel differences to sample dissimilar states and count them from successful trajectories instead of using first-visit counts in non-tabular domains. For justification of first-visit sampling, we analyzed the difference between theoretical importance, approximated importance achieved with first-visit counting, and approximated importance achieved with every-state counting of successful trajectories. In MOL, we use pseudo-count from \cite{bellemare2016unifying} for fast learning of importance.

By using first-visit pseudo-counts for approximating the importance of micro-objectives and sampling dissimilar states, we achieved significantly better results than the baselines in two Atari games: Montezuma’s Revenge and Seaquest, where we used \cite{bellemare2016unifying} as a baseline for Montezuma's Revenge, the state-of-the-art model as far as we know, and Deep Q-learning for Seaquest.

Overall, our main contributions are:
\begin{itemize}
\item Defined micro-objectives with precise measurement of their importance using visited counts combining the notions of the subgoal and the option.
\item Automatic discovery and utilization of the micro-objectives at the same time resulting in accelerated learning.
\item Micro-Objective Learning scales up to large and continuous states, such as the Atari domain.
\end{itemize}

\section{Background}
\subsection{Double Deep Q-learning}
In Double Deep Q-learning (DDQN)\cite{van2016deep}, one operator calculates the action-value and chooses the action simultaneously. This results in a positive bias. Double Deep Q-learning uses two parameters for the action-value function to solve this problem.
\begin{equation}
y_i^{DDQN} = r + \gamma \, Q(s’, \arg\max_{a’} Q(s’, a’;\theta_i);\theta^-)
\end{equation}

\subsection{Pseudo-Count Exploration}
Pseudo-Count Exploration (PSC)\cite{bellemare2016unifying} extends traditional count-based approach for efficient exploration. PSC uses a sequential density model to approximate the count of the state in large and continuous state-space. Specifically, it uses a CTS density model for each pixel to get the probability $\rho_{n}(x)$ of the state and define the recording probability $\rho’_{n}(x)$ to get the pseudo-count $\hat{N}_{n}(x)$ and pseudo-count total $\hat{n}$.
\begin{equation}
\rho_{n}(x) = \frac{\hat{N}_{n}(x)}{\hat{n}}, \;\;
\rho’_{n}(x) = \frac{\hat{N}_{n}(x)+1}{\hat{n}+1}
\end{equation}
PSC calculates the pseudo-count by solving the equations between $\rho'_{n}(x)$ and $\rho_n$.
\begin{equation}
\hat{N}_n(x) = \frac{\rho_n(x)(1-\rho’_n(x))}{\rho’_n(x)-\rho_n(x)}
\end{equation}
By pseudo-counting the states the agent has visited, it gives an additional reward to the states the agent has not seen before. 
\begin{equation}
R_{new} = R + \beta \ (\hat{N}_n(x)+0.01)^{-1/2}
\end{equation}
Direct usage of an exploration bonus results in a destabilized Q function. PSC mixed Double Q target with Monte Carlo return as below:
\begin{equation}
\begin{aligned}
\Delta Q(x_t, a_t) :=
(1-\eta)\Delta Q_{DOUBLE}(x_t, a_t) + \\ \eta[\sum_{s=0}^\infty \gamma^s((R+R_n^+)(x_{t+s}, a_{t+s}))-Q(x_t, a_t)]
\end{aligned}
\end{equation}

\section{Micro-Objective Learning}
In Micro-Objective Learning (MOL), we approximate how important a state is given the current policy and give an additional reward to the original MDP that is proportional to the importance, as below.
\begin{equation}
    R_{obj} = \alpha \cdot min(R_{max}, \frac{(1-R_{exp})}{R_{c-max}})
\end{equation}
\begin{equation}
    R_{new} = R + R_{obj}
\end{equation}
where $R_{max}$ is a constant value to limit the maximum of $R_{obj}$, $\alpha$ is a coefficient for $R_{obj}$, and $R_{c-max}$ is the current maximum value of (1-$R_{exp}$).
\begin{equation}
    R_{exp}=\frac{0.1}{\sqrt{N(s)+0.01}}
\end{equation}

$R_{exp}$ is an additional reward function used in  \cite{bellemare2016unifying} for exploration.

Let us define a successful trajectory as a trajectory that has its last state as the only goal state in the trajectory which is given in MDP. A goal state is dependent on the task, but typically we can define a goal state as a positive rewarding state. The visit-counts of the states in successful trajectories were used to estimate the importance of the states. However, counting all of the states in the successful trajectories results in poor estimation of the importance. We use dissimilar sampling, which is an extension of first-visit sampling, to solve this issue. First-visit sampling is a simple method that samples the states that have been visited for the first time when moving along the trajectory.

Giving an additional reward to the original MDP has several benefits over the option framework. By giving an additional reward, we do not need to learn the option policies, a task which is as expensive as the original RL. Also, the options need to find an initiation set and learn the termination condition to decide where to start and terminate the options, which MOL does not require. Because micro-objectives are smaller units than the subgoal or initiation set in the option framework, MOL can be viewed as a process that learns a micro-version of the option policies, but in a flexible manner.

\subsection{Empirical explanation of Micro-Objective Learning}
Intuitively, giving an additional reward to frequently visited states in successful trajectories accelerates the learning process. One possible approach to calculating importance is to count all of the states from the successful trajectories and use this count as an importance of the states. Giving an additional reward, proportional to the importance, to every state when the agent visits can encourage the agent to go to important states. However, this simple approach has two critical flaws: It creates overly-important states and it encourages the agent to revisit a certain state that gives a substantial additional reward.
\begin{figure}[t]
\vskip 0.2in
\begin{center}
\centerline{\includegraphics[width=0.7\columnwidth]{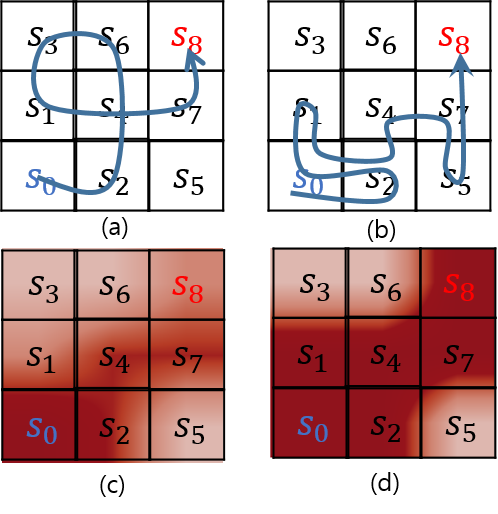}}
\caption{A simple Grid World MDP with $s_0$ as an initial state and $s_8$ as a goal, the terminal state. Figures (a) and (b) show two different successful trajectories. Figures below show the estimated importance of states using 1) every-visit counting ((c)), and 2) first-visit counting ((d)). Redder areas mean increased importance.}
\label{simple-mdp}
\end{center}
\vskip -0.2in
\end{figure}
\begin{itemize}
 \item{\bf Overly-important states}\\
Look at the simple MDP in figure \ref{simple-mdp}. Assume that the agent has succeeded in achieving the goal while exploring, as in figure \ref{simple-mdp}(c). Because the successful trajectory has a loop, states $s_1, s_3, s_6$, the states that have no relationship with achieving the goal, receive the same importance as the states $s_0, s_2, s_7$ which actually contributed to achieving the goal. If the successful trajectory has many loops, which is the normal case in early stages of exploration, the importance becomes high in the states that are included in the loops. Since we are going to give an additional reward proportional to the importance, this will result in encouraging the agent to follow the loop many times, which is definitely not what we desire.
 \item{\bf Revisiting states}\\
Let us say we have a good estimate of the importance. In the case of figure \ref{simple-mdp}(c), huge importance would be given to states $s_0, s_2, s_4, s_7, s_8$ which are the states that have contributed to earning the goal. However, if we give an additional reward to every state the agent has visited, it results in the agent going back to the states that have large importance. It is because the estimated Q value of (state, action) pairs that have states with large importance as the next state, will explode as substantial reward is continuously given. This is also not what we desire.
\end{itemize}

We solve these two problems by sampling the first-visit states from the trajectories. By using first-visit sampling of the successful trajectories before counting, we can avoid giving too much importance to any specific state because it limits the maximum count for each state in a single successful trajectory to 1. An example is shown in figure \ref{simple-mdp}. When using every-visit counting on two successful trajectories in figure \ref{simple-mdp}(c), and \ref{simple-mdp}(d), we get more importance on the states that are near the initial state because the agent has explored them frequently. However, with first-visit counting, we get a clear path to the goal state. Also, if we use first-visit sampling when we give an additional reward, the additional reward will only be given once to the state. This prevents the agent from revisiting a certain state because it will only get the reward when it first visits that state. Using first-visit sampling when giving an additional reward has one additional benefit; because we give the reward only once even if the agent visits that state several times, the expected additional reward given by visiting that state becomes $1/n$ (n is the number of times the agent has visited that certain state) times the additional reward it gets when it visits that state for the first time. We can assume the current policy of the agent has a bias for a certain state if the agent visits that state often. However, because the additional reward decreases on subsequent visits, it naturally forces the agent to go to the next state rather than forming unnecessary loops.

Defining what first-visit is in a large and continuous space is difficult because every state is different. To create the same effect as first-visit sampling, we use dissimilar sampling. Dissimilar sampling is a method that samples dissimilar states from the trajectories. This is a natural extension of first-visit sampling because the main idea of first-visit sampling is to avoid reusing the same states when calculating the importance and giving an additional rewards, and, the ‘same states’ can be extended to ‘similar states’ in large and continuous domains. To define similar states, we use the notion of pixel difference between sequential states, which has been successfully used in a 3D domain \cite{jaderberg2016reinforcement} as a criteria to find the occurrence of an event. In MOL, we use this notion as a similarity between states. While using the mean of the pixel difference in the whole trajectory seems natural, it gets awkward when the environment changes dramatically. This pixel difference $\delta$ can be viewed as
\begin{equation}
\delta = \delta_{agent} + \delta_{env}
\end{equation}

Though we need to take the environment into account, the environment alone may have too great an effect to be a good criterion for sampling. This can result in sampling only the states that have huge environmental changes, even if they are not related to the task. To address this issue, we use the mean of recent pixel differences as a criteria because recent pixel difference will exclude meaningless changes in the current environment.

While algorithm \ref{alg:dissimilar sampling} takes the trajectory as an input which assumes using an off-policy algorithm, we can still do dissimilar sampling when using an on-policy algorithm because we sample the states in sequential order. At each step, we can update the recent history of the states and decide whether the current state should be chosen or not. If the state is chosen, we add the state into the sampled trajectory and give an additional reward to it. When the reward is given, we update the pseudo-count density model with the sampled trajectory and re-initialize the sampled trajectory. Remember, a successful trajectory is defined as a trajectory that ends with the only goal state in the trajectory. This means that there can be several successful trajectories in a single episode. Splitting an episode trajectory into several successful trajectories is reasonable because for each reward, micro-objectives should be counted separately. In other words, if a single state is important in acquiring several goals, we need to count it several times to represent its importance.

\begin{algorithm}[tb]
   \caption{Dissimilar Sampling}
   \label{alg:dissimilar sampling}
\begin{algorithmic}
   \STATE {\bfseries Input:} A trajectory $L =$ ($s_0, s_1, … , s_n$), RecentHistorySize = $h$
   \STATE {\bfseries Output:} sampled trajectory $L^*$
   \STATE $L^*$ = [$s_0$]
   \FOR{$i=1$ {\bfseries to} $n$}
   \STATE $\delta$ = $||((s_{i-h}:s_i)-(s_{i-h+1}:s_{i+1}))||$
   \IF{all($||(L^*[j]-s_i)||$ $\geq$ $\delta$) (for $j = 1$ to the length\\ of ($L^*$)))}
   \STATE Append $s_i$ to $L^*$
   \ENDIF
   \ENDFOR
\end{algorithmic}
\end{algorithm}

When designing an additional reward function, we have considered several requirements.
\begin{itemize}
  \item {\bf Convergence}
It must have limits as the pseudo-count grows larger. Though we are going to clip the reward, an expanding reward is undesirable because it makes the Q function unstable.
 \item {\bf Early stage exploitation}
 It must be significant, even in the early stages of learning, because exploiting micro-objectives is most effective when the learning is more imperfect.
 \item {\bf Distinguish micro-objectives and non micro-objectives}
 It must be able to distinguish between micro-objectives and non micro-objectives, which have an importance of nearly 0.
\end{itemize}

Scaling the reward with a maximum value of (1-$R_{exp}$) for steps leading to the current learning step, will result in giving substantial reward values in the early stages of learning. Also, we clip the reward with $R_{max}$ to limit the maximum additional reward.

\begin{algorithm}[tb]
   \caption{Micro-Objective Learning in Deep Q-Learning}
   \label{alg:micro-objective learning}
\begin{algorithmic}
   \STATE Initialize replay memory $D$ and action-value function $Q$
   \STATE Initialize trajectory $L$ and pseudo-count model $M$ for $R_{obj}$
   \FOR{episode=$1$ {\bfseries to} $n$}
   \REPEAT
   \STATE Select an action $a$ with epsilon-greedy policy
   \STATE Do action $a$ in current state $s$
   \STATE Update parameters in function $Q$ using replay memory $D$
   \STATE Use dissimilar sampling($L+s', h$) to get sampled trajectory $L^*$ where $s'$ is the next state
   \IF{$s'$ $\in$ $L^*$}
   \STATE $R$ = $R$ + $R_{obj}(s')$
   \STATE Append $s'$ to $L$
   \ENDIF
   \STATE Insert $(s, a, s', R, terminal)$ in replay memory $D$
   \IF{next state is a goal state(external reward $> 0$)}
   \STATE Update $M$ with $L$
   \STATE Re-initialize trajectory $L$
   \ENDIF
   \UNTIL{Terminal is True}
   \ENDFOR
\end{algorithmic}
\end{algorithm}

\subsection{Analysis on Micro-Objective Learning}
Consider a Markov Decision Process (MDP) which is defined with $(S, A, R, \rho_0, \gamma, P)$. $S$ is a set of states, $A$ is a set of possible actions, $R$ is an external reward from the environment, $\rho_0$ is an initial state distribution, gamma is a discount factor, and $P : S \times A \times S \rightarrow \mathbb{R}$, is the transition probability distribution.

We define the importance of a state in a given trajectory.

\theoremstyle{definition}
\begin{definition}{Importance Count}\\
Let there be a successful trajectory $L$ with visited states $s_0$ to $s_n$ in sequential order. We define a {\bf importance count} $I^L(s_i)$ for every state $s_i$ in a MDP as follows, where $L^*$ is the optimal path from $s_0$ to $s_n$:
% \begin{center}
% $I^L (s_i)$ = 
% \begin{cases}
%  & 1 \text{ if } $s_i$ $\in$ $L^*$  \\ 
%  & 0 \text{ otherwise}
% \end{cases}
% \end{center}
\end{definition}

\begin{equation*}
I^L (s_i) = 
\begin{cases}
 & 1 \text{ if } s_i \in L^*  \\ 
 & 0 \text{ otherwise}
\end{cases}
\end{equation*}

\begin{figure}[t]
\vskip 0.2in
\begin{center}
\centerline{\includegraphics[width=\columnwidth]{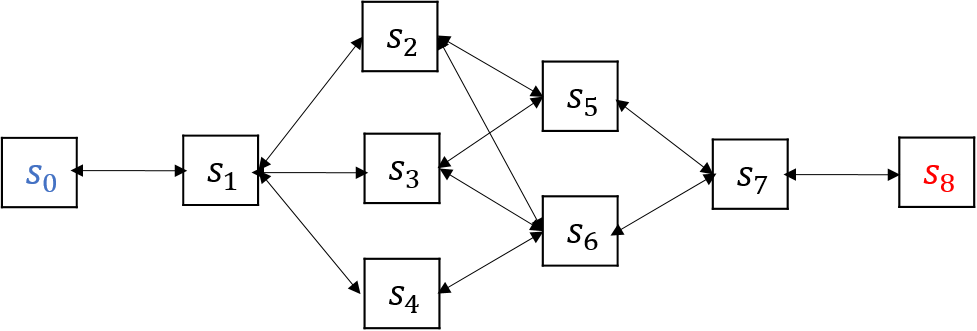}}
\caption{A simple MDP with $s_0$ as an initial state and $s_8$ as a goal state.}
\label{simple-mdp2}
\end{center}
\vskip -0.2in
\end{figure}

For example, in the MDP shown in figure \ref{simple-mdp2}, assume the successful trajectory $L$ has all (state, next state) pairs except for $(s_6, s_7)$. By definition, we get importance of 1 except for the states $s_4$, and $s_6$. These two states did not contribute to reaching the goal state. This definition naturally arises from what humans do when searching for valuable steps in a complex task. Humans try to figure out why a trial was successful by tracing back the cause of the success, which is a similar process to finding the optimal path given a successful trajectory. We use the importance count to define the importance of a state given the policy.

\begin{definition}{Micro-objective} \\
A micro-objective is a state $s_i$ that has importance $M_{\pi}(s_i)$ $> 0$ given the current policy $\pi$. Let $H$ be the set of possible successful trajectories in the given MDP.
\begin{equation}
M_{\pi}(s_i) = \sum_{L \in H} I^L(s_i) \cdot p_{\pi}(L)
\end{equation}
, where $p_{\pi}(L)$ is the probability of following the trajectory $L
$ when using the policy $\pi$.
\end{definition}
The state importance should be dependent on the current policy as the value function $V(s)$. For example, in the MDP in figure \ref{simple-mdp}, if we have a policy of going up, then the important states would be the states from below because in the current policy, getting to the states in the bottom row makes the probability of getting to the goal state higher. However, if we have a policy of going down, the states that are above are more important for reaching the goal state.

Considering that the definition of importance count is whether or not the state has contributed to achieving the goal, the importance of the micro-objective defines how likely we succeed if we are in that state using the current policy. Because we update the policy at every step, using recent successful trajectories is a reasonable approximation. For convenience, we used all successful trajectories to estimate the importance of micro-objectives.

We argue that giving an additional reward proportional to the estimated importance accelerates learning to reach the goal. Because of the discount factor $\gamma$, states near the initial states have small $V(s)$. Also, to update the $V(s)$ of those states, updating $V(s)$ of the states between those states and the goal state is required. However, if we give an additional reward to each state, $V(s)$ will be updated quickly and the requirements mentioned above are eliminated. For example, in the figure \ref{simple-mdp2} MDP, to update the value of state $s_1$, the values of state $s_k$ $(k \geq 2)$ need to be updated. However, when giving an additional reward, a direct update is possible with $(s_1, a, s_j)$ $(j = 2, 3, 4)$ in the replay pool $D$.
\begin{figure}[t]
\vskip 0.2in
\begin{center}
\centerline{\includegraphics[width=\columnwidth]{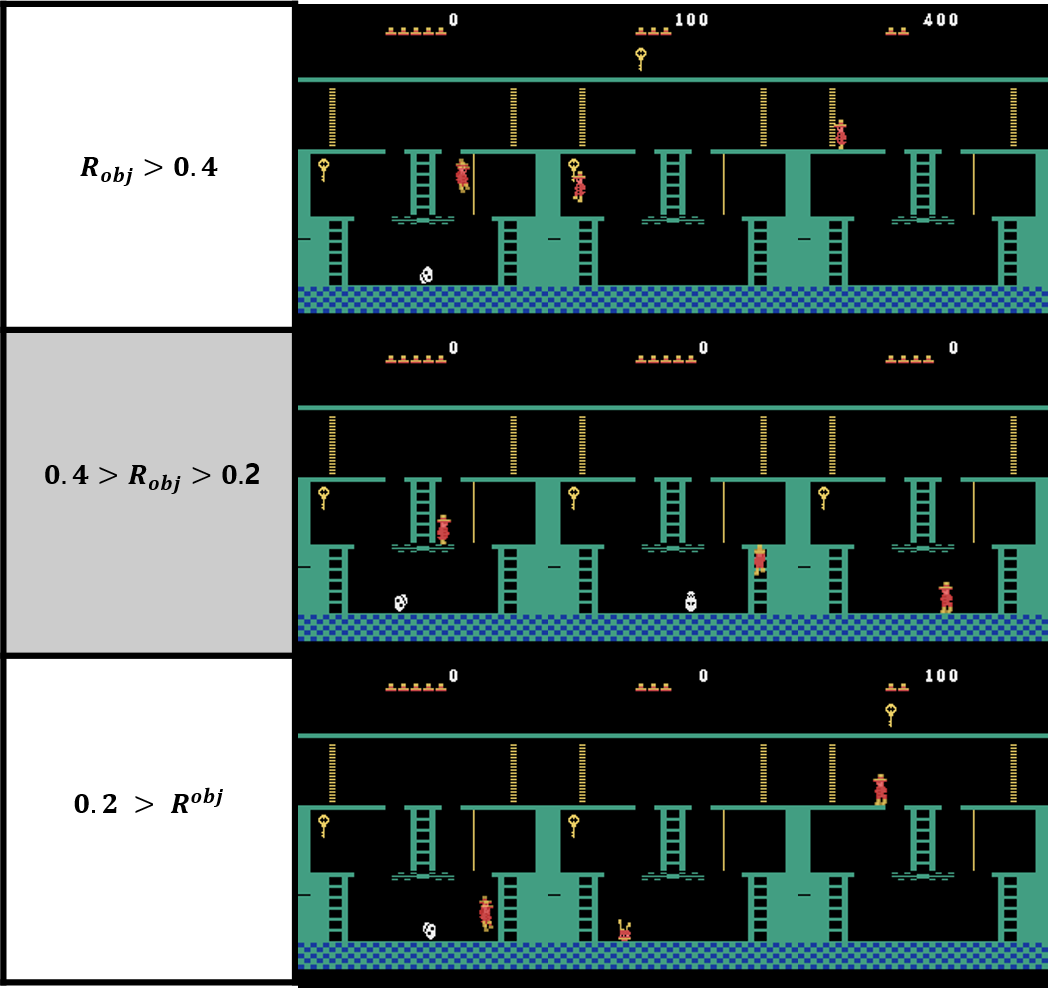}}
\caption{The states that have the largest, medium, and smallest $R_{obj}$ in Montezuma's Revenge: First, second, and third row each. For Montezuma's Revenge, we used 0.2 and 0.4 as the criteria. The states with the largest $R_{obj}$ are traditional subgoal states or actual goal states (key, rope, and the door). The states with the medium $R_{obj}$ are not necessary to reach the goal states, but are helpful. The states with the smallest $R_{obj}$ do not have much relationship with the goal states.}
\label{montezuma-frames}
\end{center}
\vskip -0.2in
\end{figure}
While the benefits of giving an additional reward to the appropriate states are obvious, estimating the appropriate states is not. When using first-visit sampling, we can effectively approximate the state importance. Assuming that we are using a substantial number of successful trajectories obtained from the current policy, with an estimated importance count $I^L_{est}$, actual importance count $I^L$, the set of states $S$, and the difference between the estimated importance $M_{est}$ and the actual importance $M$, $L_{M}$,
\begin{equation}
\label{equation_loss}
L_{M_{est}} = \sum_{s_i \in S}\sum_{L \in H} (I^L(s_i)-I^L_{est}(s_i)) \cdot p_{\pi}(L)
\end{equation}
As in equation \ref{equation_loss}, the loss comes from the difference of importance count, which is caused by states that are not included in the optimal path of a successful trajectory.

To approximate the loss of importance count $(I^L(s_i)-I^L_{est}(s_i)$, we analyze how the agent is trained. When giving rewards to every state that is visited in a successful trajectory, there can be states that distract the agent from reaching the goal.  With appropriate exploration methods, the count of these states will be low. Also, though the agent may not follow the optimal trajectory, since the estimated importance is as follows,

\begin{equation}
M_{est}(s_i) = \sum_{L \in H} I^L_{est}(s_i) \cdot p_{\pi}(L)
\end{equation}
the agent is encouraged to follow the successful trajectory that is the most likely in the current policy, resulting in fast convergence of the policy for getting to the goal state. 
\begin{figure}[t]
\vskip 0.2in
\begin{center}
\centerline{\includegraphics[width=\columnwidth]{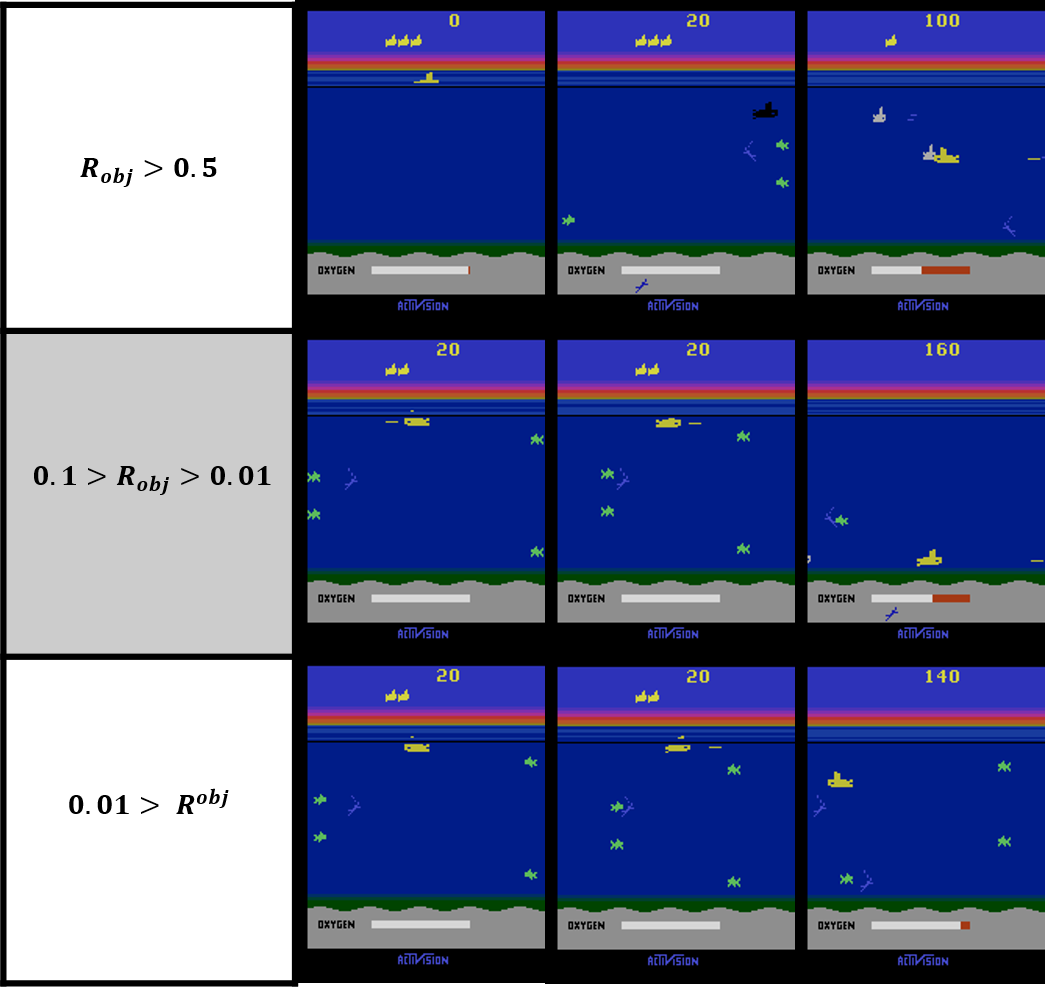}}
\caption{The states that have the largest, medium, and smallest $R_{obj}$ in Seaquest: First, second, and third row each. For Seaquest, we used $0.01, 0.1$, and $0.5$ as the criteria. The states with the largest $R_{obj}$ are states that give an external reward. The states with the medium $R_{obj}$ are the states where the submarine is firing its weapon, which is a way to get an external reward. The states with the smallest $R_{obj}$ do not have much relationship with the goal states.}
\label{seaquest-frames}
\end{center}
\vskip -0.2in
\end{figure}
Therefore, though estimated convergence by first-visit sampling does not converge to the actual importance, $L_{M_{est}}$ converges. Though MOL does not guarantee an optimal policy, it helps for an agent to learn the policy that succeeds in reaching the goal state and this makes it possible to get to the next reward, resulting in a higher average score.

\section{Experiments}
In the experiments, we focused on 1) analyzing which states are chosen as micro-objectives and which states have large or small importance, and 2) evaluating how much MOL accelerates learning using the discovered micro-objectives. We compared our agent to existing methods in two Atari games: Montezuma's Revenge and Seaquest. Montezuma’s Revenge is notorious for its difficult exploration while Seaquest is one of the most dense reward games in Atari.

In Montezuma's Revenge, we compared a pseudo-count exploration model from \cite{bellemare2016unifying}, which is the state-of-the-art model in this domain, with and without MOL. This was because Deep Q-learning has a difficult time obtaining successful trajectories which are needed for MOL. We tested our agent in a stochastic ALE setting with a probability to repeat previous actions of 0.25, the same setting as in \cite{bellemare2016unifying}. 
\begin{figure}[ht]
\vskip 0.2in
\begin{center}
\centerline{\includegraphics[width=\columnwidth]{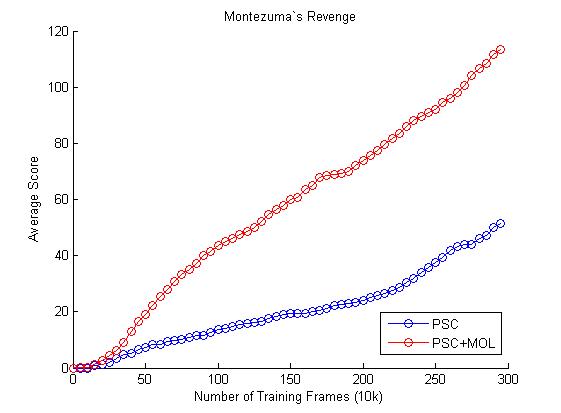}}
\caption{Average Training Episode Score of Montezuma's Revenge for 3 million training frames with three random seeds - Pseudo count exploration model from \cite{bellemare2016unifying} versus MOL with Pseudo-count exploration model.}
\label{montezuma-result}
\end{center}
\vskip -0.2in
\end{figure} 
In Seaquest, we compared the Double Q-learning with Monte-Carlo return, with and without MOL. Gym environment was used as an emulator\cite{1606.01540}. Parameters used are set the same as \cite{van2016deep}.

For dissimilar sampling, we used recent history size h = 5, and 2,500 as a minimum pixel difference for clear sampling. We reset the emulator every 250,000 frames to lower the memory usage as extremely long episodes take too much memory. Also, we used $\alpha = 1$ and $R_{max} = 0.9$ for $R_{obj}$.

We trained the agents for 3 million frames in both games with three different random seeds. In Montezuma's Revenge, after getting the reward of 300 by reaching the door, the rewards following that come from exploring additional rooms. 3 million frames were chosen to see how the agent is trained in diverse rooms. Also, in Seaquest, it is sufficient to observe the effect of MOL.

\subsection{Analysis on Micro-Objectives}

To analyze the pseudo-count used for $R_{obj}$, we took one successful trajectory of the agent trained with 3 million frames. We sampled with dissimilar sampling and compared $R_{obj}$ of the sampled frames. Figures \ref{montezuma-frames} and \ref{seaquest-frames} show three sample states with the largest, medium, and the smallest $R_{obj}$ for each game when training on 3 million frames.

As we can see in the figure \ref{montezuma-frames}, in Montezuma's Revenge, the states where the character is reaching the key, rope, and the door have the largest counts, which are traditional subgoal states. 
\begin{figure}[ht]
\vskip 0.2in
\begin{center}
\centerline{\includegraphics[width=\columnwidth]{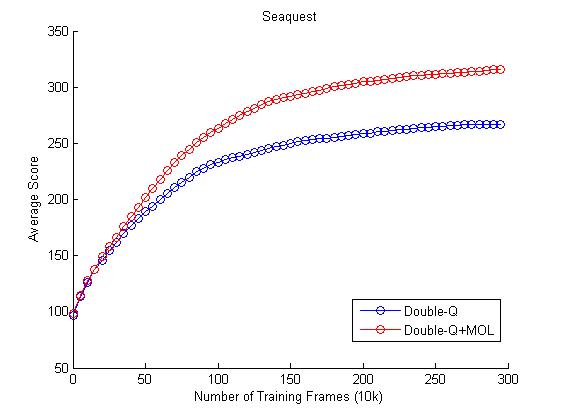}}
\caption{Average Training Episode Score of Seaquest for 3 million training frames with three random seeds - Double Q-learning with Monte Carlo return(a model from \cite{bellemare2016unifying} without exploration bonus) versus MOL with Double Q-learning with Monte Carlo return model.}
\label{seaquest-result}
\end{center}
\vskip -0.2in
\end{figure}
However, the states which have medium counts are not traditional subgoal states, but are important in reaching the key or the door. In the first sample, the character is reaching the rope while he does not need to use that path to reach it. Also in the third sample, the character does not need to go left since he can also jump to the left. The states which have the smallest counts appear to have weak relationships with the goal states.

In Seaquest, the initial game states and goal states have the largest counts as in figure \ref{seaquest-frames}. In the states with medium counts, the submarine appears to do the "fire" action which is needed to get additional points. As in Montezuma's Revenge, the states which have the smallest counts seem to have weak relationships with the goal states.

\subsection{Accelerating Learning}
The learning curve of both games are shown in figure \ref{montezuma-result} and \ref{seaquest-result}. We averaged the training episode scores of three experiments. In Montezuma's Revenge, because the reward is sparse and the subgoal state is clear, the gap dramatically increases as expected. After three million frames of training, the average training episode score of the agent with MOL exceeded 100, which means the agents are constantly getting to the door after obtaining the key. Meanwhile, Seaquest is a dense reward game which is rather hard to interpret the subgoal states. However, even in Seaquest, the gap between the baseline and MOL increased as training proceeds. This suggests MOL can be applied to games with unclear subgoal states. After training 3 million frames, using MOL resulted in 120.34\%, and 18.25\% increase in Montezuma's Revenge and Seaquest scores, respectively.
\begin{table}[t]
\label{ratio-table}
\vskip 0.15in
\begin{center}
\begin{small}
\begin{sc}
\begin{tabular}{lcc}
\hline
\abovespace\belowspace
Algorithm & Seaquest & Montezuma's\ Revenge \\
\hline
\abovespace
Baseline       & 267.10 $\pm$\ 11.88 & 51.40 $\pm$\ 10.73 \\
With MOL       & 315.84 $\pm$\ 15.01 & 113.26 $\pm$\ 40.62\\
\hline
\abovespace
Ratio(\%) & 18.25 & 120.34 \\
\hline
\belowspace
\end{tabular}
\end{sc}
\end{small}
\end{center}
\vskip -0.1in
\caption{Comparison between the baseline and with MOL after training on 3 million frames.}
\end{table}

\begin{figure}[ht]
\vskip 0.2in
\begin{center}
\centerline{\includegraphics[width=\columnwidth]{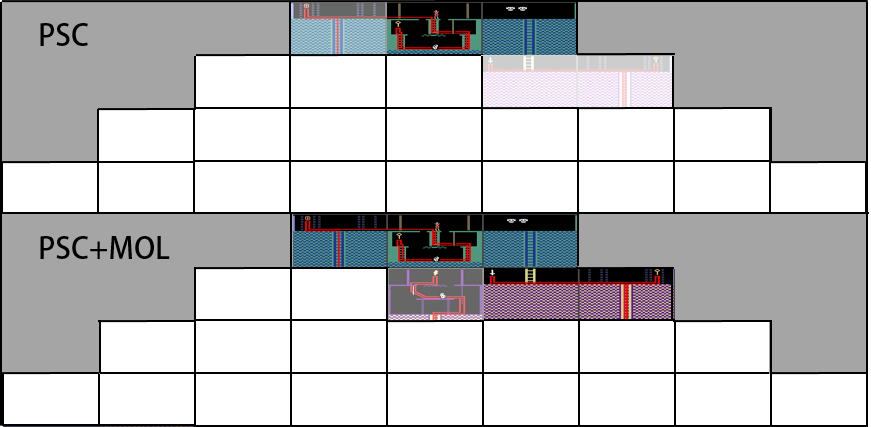}}
\caption{Explored rooms in Montezuma's Revenge after 1.5 million training frames - PSC only explored room 1 and 2 in all 3 experiments with room 0 explored in 2 experiments and room 6 and 7 explored in only 1 experiment. PSC+MOL explored room 0, 1, 2, 6, and 7 in all 3 experiments with room 5 explored in 2 experiments.}
\label{montezuma-room}
\end{center}
\vskip -0.2in
\end{figure}

Using MOL can be viewed as an exploitation of the successful trajectories which might distract exploration. However, the agent can explore better with improved exploitation methods. As in figure \ref{montezuma-room}, with MOL, the agent normally explores 6 rooms (one experiment fails to explore room 5), before 1.5 million training frames. The fastest room search was exploring 6 rooms in 0.4 million training frames while the baseline searches 6 rooms after 5 million training frames according to \cite{bellemare2016unifying}. Also, in one experiment, the agent with MOL started to get the external reward of 2,500 before training 1.5 million frames, while the agent needs to collect three items (a key, a door, and a knife) to reach a score of 2,500.

\section{Conclusion}
In this paper, we proposed an autonomous and effective hierarchical reinforcement learning method, Micro-Objective Learning, which accelerates learning by setting micro-objectives with pseudo-counts. Using dissimilar sampling, we avoided counting and giving rewards to similar states, which was critical to discovering precise micro-objectives and learning. Experimental results in Montezuma’s Revenge and Seaquest show that micro-objectives embrace the subgoals which were heuristically designed previously and are effective in both sparse and dense reward settings.

In this work, we have successfully applied the notion of pixel difference for dissimilar sampling. However, for generalization, we have to use higher level features instead of pixel difference. Currently, we need to pre-train the networks to get higher level features, but it does not give sufficiently good features. With additional exploration in unsupervised learning, we could generalize further. In addition, although giving an additional reward directly to an original MDP is effective in accelerating the learning process, it does not guarantee convergence to the optimal policy. Therefore, finding a way to guarantee convergence while still getting the advantages of directly giving an additional reward will be our next step.

\section*{Acknowledgements} 
The authors would like to thank Heidi Lynn Tessmer for discussion and helpful comments.
\nocite{silver2016mastering}
\nocite{sutton1999between}
\nocite{konidaris2009skill}
\nocite{bellemare2012investigating}
\nocite{stolle2002learning}
\nocite{stadie2015incentivizing}
\nocite{hasselt2010double}
\nocite{jaderberg2016reinforcement}
\nocite{chiu2011subgoal}
\nocite{csimcsek2005identifying}
\nocite{goel2003subgoal}
\nocite{bakker2004hierarchical}
\nocite{oh2015action}
\nocite{houthooft2016vime}

\bibliography{example_paper}
\bibliographystyle{icml2017}

\end{document}